\documentclass{article}
\usepackage{amssymb}
\usepackage{amsmath}
\usepackage{rotating}
\usepackage{graphicx}
\usepackage{float}
\usepackage{multicol}
\usepackage[bottom]{footmisc}
\usepackage{newtxtext}       %
\usepackage{newtxmath}       
\usepackage{hyperref}\usepackage{breakurl} 
\usepackage{algorithm}
\usepackage{algorithmic}
\usepackage{subfigure}
\makeindex             
\newcommand{\ma}[1]{\ensuremath{\mathbf{#1}}}
\newcommand{\R}{\mathbb{R}}

\newcommand{\mbf}[1]{\mathbf{#1}}
\newcommand{\abs}[1]{{\left\vert  #1 \right\vert}}
\newcommand{\norma}[1]{\left\|   #1 \right\|}
\newcommand{\dsum}{\displaystyle \sum}
\newcommand{\sistema}[1]{{\left\lbrace  \begin{array}{lll}#1 \end{array}   \right. }}
\newcommand{\Contador}{\text{\footnotesize \textsf{Counter}} }

\begin{document}

\title{Clustering via Ant Colonies: Parameter Analysis and Improvement of the Algorithm}
\author{Jeffry Chavarr\'ia-Molina\thanks{School of Mathematics, Costa Rica Institute of Technology, Cartago, Costa Rica. \url{jfallas@itcr.ac.cr}} \and 
Juan Jos\'e Fallas-Monge\thanks{School of Mathematics, Costa Rica Institute of Technology, Cartago, Costa Rica. \url{jchavarria@itcr.ac.cr}} \and 
Javier Trejos-Zelaya\thanks{CIMPA--School of Mathematics, University of Costa Rica, San Jos\'e, Costa Rica. \url{javier.trejos@ucr.ac.cr}}}

\maketitle

\begin{abstract}
An ant colony optimization approach for partitioning a set of objects is proposed. In order to minimize the intra-variance, or within sum-of-squares, of the partitioned classes, we construct ant-like solutions by a constructive approach that selects objects to be put in a class with a probability that depends on the distance between the object and the centroid of the class (visibility) and the pheromone trail; the latter depends on the class memberships that have been defined along the iterations. The procedure is improved with the application of K-means algorithm in some iterations of the ant colony method. We performed a simulation study in order to evaluate the method with a Monte Carlo experiment that controls some sensitive parameters of the clustering problem. After some tuning of the parameters, the method has also been applied to some benchmark real-data sets. Encouraging results were obtained in nearly all cases.
\end{abstract}

\noindent
\textbf{Keywords}: clustering; ACO; machine learning; ant colony optimization; intraclass variance; TSP; heuristics; algorithm; simulation.

\noindent
\textbf{Mathematics Subject Classification}: 91C20, 62H30, 90C59.

\section{Introduction}
\label{sec:1}

Cluster analysis, or clustering, is one of the main tools in Data Analysis and Machine Learning, since it intends to discover groups or classes in large data sets of objects described by observed variables, simplifying this way the set with a small number of clusters. 
Most clustering methods are based on dissimilarities, graphs, models or densities. In our case, we will deal with dissimilarities or distances for numerical data sets.
There are two main families in this case: partitioning methods and hierarchical ones, being K-means and agglomerative hierarchical methods, respectively, the most widely used in practice. Both have local optimality problems: local minima that depend on initialization for K-means, greedy procedure for agglomerative hierarchical clustering.

Several combinatorial optimization metaheuristics have been used for cluster partitioning \cite{Handl:Knowles,K+Wong,Sarkar+Yegnanarayana,Trejos:1998}. In this article we deal with partitioning for numerical data sets, using an ant colony optimization (ACO) approach in order to overcome the local optima problem.

According to \cite{Handl:Knowles}, published in 2006, ``a few implementations of ACO have been proposed for data-clustering, with the construction graph typically employed to directly represent cluster assignments \cite{Handl:Meyer,Runkler}´´.

In 2004, we published a first paper on clustering using an ant colony optimization approach \cite{Trejos+Murillo+Piza+CACO} for the minimization of the within sum-of-squares criterion. In that method, ants were associated with partitions that were modified during the iterations, according to a probability of selection that depends on the visibility (proportional to the distance between the objects) and the pheromone trail (which depends on the fact that the objects have been classified together in the partitions). 
The pheromone matrix measured relation intensity between pairs of objects.

By that time, Shelokar, Jayaraman \& Kulkarni \cite{Shelokar} published another clustering method based on ACO for minimizing the same criterion as in \cite{Trejos+Murillo+Piza+CACO}, with a pheromone trail but no local heuristic.
The pheromone matrix relates objects and clusters, and it is defined by the inverse of the objective function.
The matrix is used as a kind of adaptive memory that contains information provided by the previously found superior solutions, and is updated at the end of each iteration \cite{Shelokar}. This information is considered by the other ants to continue the clustering process.
However, it is not clear how the authors selected the parameters to execute the ACO algorithm. They indicate that several simulations were performed to find the algorithm parameters \cite{Shelokar}, but they do not present details about the process. 
They also present a comparison among their ants algorithm and other heuristic methods such as genetic algorithm, simulated annealing and tabu search. 

Later on, Kao \& Cheng (2006) in a short paper \cite{Kao:Cheng} improved Shelokar's algorithm introducing a local heuristic or visibility based on the inverse of the distance between objects and class centers. The pheromone trail is also defined by the inverse of the criterion and the algorithm follows almost the same steps as Shelokar algorithm \cite{Shelokar}, with the difference that visibility is introduced.

Neither \cite{Shelokar} nor \cite{Kao:Cheng} give a detailed analysis on the choice of parameters for their methods.


In the present article, we use ACO with ants constructing  partitions. 
The strategy is based on the traveling salesman problem (TSP) in a similar way as it was tackled in \cite{Bonabeau} with ACO, in our case for the clustering problem. 
It is a constructive method, in which each ant builds a partition. 
This part of the process is similar to the ideas presented in \cite{Kao:Cheng} and \cite{Shelokar}, which were previously presented; but this paper deals with three different aims: first, developing a fitting parameters analysis 
studying the algorithm behavior in the clustering problem according to its parameters. Second, we introduce a local search procedure based on the K-means algorithm, to improve the basic ACO (BACO) algorithm performance. And finally, to develop a performance comparison among the K-means algorithm (KM), the BACO algorithm and the BACOK (BACO improved with the local search procedure) algorithm.     

The article is organized as follows. Section \ref{sec:2} contains the mains concepts of clustering we use in the article, introducing the main notation we need.
In Section \ref{SeccionColoniaHormigasArtif} the artificial ant concept is explained and the ACO classical algorithm is presented. In Section \ref{SeccionACOParaClustering} we introduce the proposed ACO algorithm. Section \ref{SeccionExperimental} describes the experiment performed. Sections \ref{SeccionDiscusionResultados} and  \ref{SeccionConclusiones} present the results and some remarks.

\section{Clustering}
\label{sec:2}

Cluster analysis, or clustering, deals with finding homogeneous groups of objects such that similar objects belong to the same class and it is possible to distinguish between objects in different classes. Cluster analysis can be defined as an optimization problem in which a given function consisting of \textit{within cluster similitary} and \textit{among clusters dissimilarities} need to be optimized \cite{Jafar:Sivakumar,Adilson}. In the numerical case, there is a set of objects $\Omega = \{\ma{x}_1,\ma{x}_2,\ldots,\ma{x}_n\}$ such that
$\ma{x}_i  \in \R^p$, for all $i$, that is, the objects are described by $p$ numerical or quantitative variables. 
The most widely used criterion \cite{Everitt} is the minimization of the within sum-of-squares, also known as within inertia or variance:
\begin{equation}
W = \frac 1n \sum_{k=1}^K \sum_{\ma{x}_i\in C_k} \|\ma{x}_i - \ma{g}_k \|^2,
\label{W}
\end{equation}
where $K$ is the number of classes or clusters (number fixed a priori), $P = (C_1,C_2,\ldots,C_K)$ is a partition of $\Omega$, and $\ma{g}_k$ is the barycenter or mean vector of $C_k$. 
Minimizing $W(P)$ is equivalent to maximizing the between sum-of-squares (between inertia or variance):
\begin{equation*}
B = \sum_{k=1}^K \frac{|C_k|}{n} \|\ma{g}_k - \ma{g} \|^2,
\label{B}
\end{equation*}
where $\ma{g}$ is the overall barycenter and $|C_k|$ is the cardinality of class $C_k$, 
since the sum $I = W(P)+B(P)$ is a constant (the total inertia) \cite{Everitt}.

The $W(P)$ function is not a convex function, thus $W(P)$ could have several local minima \cite{K+Wong,Sarkar+Yegnanarayana}. This feature causes the traditional clustering algorithms based on local search, such as K-means, to find mostly local minima \cite{Trejos:1998}. Furthermore, the global optimization algorithms (such as linear programming, interval methods, branch and bound methods, etc.) present a high sensitivity to relatively high dimensional data tables, in which the algorithms' probability for finding the optimal partition is very low. In those cases, algorithms report solutions that differ significantly from the optimum clustering \cite{Bagirov:Mardaneh}. Those features represent a challenge to try to find alternative optimization strategies, and combinatorial optimization heuristics are a viable option. 

In recent years heuristic algorithms have been used to solve complex optimization problems, since their random nature is useful to efficiently avoid the convergence to local minima \cite{Babu:Murty,Klein:Dubes,Trejos:1998}. As particular examples of optimization heuristics used in clustering it is possible to cite simulated annealing, tabu search, genetic algorithms, particle swarm optimization and ant colony optimization.  

In the particular case of ant colony optimization, there are several contributions, as the already mentioned \cite{Kao:Cheng,Shelokar,Trejos+Murillo+Piza+CACO}, and some other more recent \cite{Handl:Meyer,Handl:Knowles,Runkler,Zhe}.
 
\section{Artificial ant colonies}
\label{SeccionColoniaHormigasArtif}

The optimization approach based on ant colonies (ACO) is part of a large group based on swarm intelligence. It was proposed by Marco Dorigo in 1992, to solve several discrete optimization problems \cite{Dorigo+DiCaro+Gambardella, Jafar:Sivakumar}, and since then it has been applied to several combinatorial optimization problems. 
This method, like every metaheuristic, depends on parameters which control several decisions taken in the process. There are several papers which develop parameters analysis for the ACO algorithm. In \cite{Gaertner+Clark} an empirical analysis
of the sensitivity of the ACO algorithm to variations of some parameters for different instances of the TSP (traveling salesman problem) is presented. Similarly, in \cite{Wei} an experiment with parameter combinations is shown, in order to improve the speed of convergence of the ACO algorithm in the TSP. Also, this author indicates that at present the parameter settings and properties research of basic ant colony algorithm are mostly still in the experimental stage \cite{Wei}. Meanwhile, \cite{Stutzle+Dorigo} provides an extensive review of available research results on parameter adaptation in ACO
algorithms. They mention that ACO algorithms involve a number of parameters that need to be set appropriately, in particular $\alpha$, $\beta$ (both used to weigh the relative influence of the pheromone) and $\rho$ (evaporation rate parameter, $0\leq \rho \leq 1$). A parameter selection in the TSP context is developed in \cite{Dorigo+Maniezzo+Colorni}, in three different experiments. They tested the ranges: $\alpha\in \{0,0.5,1,2,5\}$, $\beta\in \{0,1,2,5\}$, $\rho \in \{0.3, 0.5, 0.7, 0.99, 0.999\}$ and $Q\in \{1,100,10000\}$. The numbers $\alpha=1$ and  $\beta=5$, were selected as the best values for this parameters. Parameter $\rho$ was fixed, depending on the experiment, in $0.99$, $0.99$ or $\rho=0.5$. And finally, parameter $Q$ was found to be negligible. 

In nature, the optimization developed by ants while they look for food consists basically of minimizing the distance between the nest and food. For this reason the first application of ACO was to the TSP \cite{Bonabeau}. In that problem the agent should visit $n$ cities, all interconnected, visiting all cities just one time and then returning to the departure city, minimizing the distance.      

In this paper the TSP idea is used to study the clustering optimization problem. Thus, it is necessary to introduce artificial ants; that is, agents in charge of finding a feasible solution in the search space. During this proccess the ant will drop artificial pheromones so that other ants can rebuild the same solution. Pheromones should be volatile (disappear in time on the trails that have not been intensified) and have to increase on the shortest trails while the number of iterations increases \cite{Dorigo+DiCaro+Gambardella}.    

The pheromone update formula applied in the TSP is given by  $ \tau_{uv}=(1-\rho) \tau_{uv}+\rho\Delta \tau_{uv} $ \cite{Barcos+Rodriguez+Alvarez+Robuste,Dorigo+Birattari+Stutzle,Dorigo+Gambardella+ACSCLATSP}, where $ \tau_{uv} $ is the pheromone present on the trail from $ u $ to $ v $, $ \rho $ is the evaporation rate, and 
\[ \Delta \tau _{uv}=\sum_{m=1}^{M}\Delta \tau_{uv}^{m}, \]
where $ M $ is the number of ants, and $ \Delta\tau_{uv}^{m} $ is the pheromone dropped by the $m$-th ant on the trail $ (u,v) $, normally given by:
\[ \Delta\tau_{uv}^{m}=\begin{cases}
Q/d_{m} & \text{ if ant } m \text{ walks across } (u,v) \\
0& \text{ otherwise;}
\end{cases} \]
where $Q$ is a parameter to be fitted and $ d_{m} $ represents the total distance walked by ant $m$.

An alternative way to deal with pheromones is to make local updatings, that is, every time an ant goes from node $u$ to node $v$, a local pheromone update is applied on the trail $(u,v)$ \cite{Dorigo+Gambardella+ACSCLATSP}. A possible local update formula is $ \tau_{uv}=\tau_{uv}+\dfrac{Q}{d_{uv}},$
where $Q$ is a parameter to be fitted and $ d_{uv} $ is the distance between $u$ and $v$. When all ants finish their trips, the pheromone is updated by applying the evaporation rate.

On the other hand, each ant has to decide to which node it goes from the current node. In that choice three factors are fundamental: visibility, pheromone trail and a probabilistic factor. Thus, if $ T_{m} $ represents the route built by the ant $m$ while it is on the node $u$, then the probability of going to the node $v$ is given by:
\[ p_{uv}^{m}=\begin{cases}
\dfrac{[\tau_{uv}]^{\alpha} \cdot [\eta_{uv}]^{\beta} }{ \sum\limits_{s\not \in T_{m}} [\tau_{us} ]^{\alpha} [\eta_{us}]^{\beta}} & \text{ if }v \not\in  T_{m} \\
0 & \text{ if } v\in T_{m};
\end{cases} \]
where $ \eta_{uv} $ is the visibility, defined by $ \eta_{uv}=1/d_{uv} $, with $ d_{uv} $ the distance from the node $ u $ to node $ v $; $ \tau_{uv} $ is the pheromone on the trail $(u,v)$, and $ \alpha $ and $ \beta $ are parameters to be fitted \cite{Barcos+Rodriguez+Alvarez+Robuste,Dorigo+DiCaro+Gambardella,Kennedy+Eberhart}.

To stop the algorithm, \cite{Bonabeau} proposed using a maximum iteration number. The disadvantage of this procedure is that it could stop the algorithm while it is still improving the solutions. Also, \cite{Dorigo+Maniezzo+Colorni} considered investigating a stagnation behavior of all ants traveling the same path. A stagnation process is present if a percentage of the ants have the same distance in their paths. Thus, it is almost certain that those ants are traveling the same path, or at least, that they are traveling paths with the same cost value.

In algorithm \ref{alg:Hormigas}, the classical ACO algorithm is shown.

\begin{algorithm}
\begin{algorithmic}[1]
\REQUIRE 	Initial parameters.
\STATE Set parameters and initialize pheromone trails. 
\WHILE  {stop criterion is not satisfied}        
	\FOR{$ t\leftarrow1 $ \TO total of nodes}
		\FOR{ $ m \leftarrow1 $ \TO $ M $}
			\STATE Move ant $m$ to a new position.
			\STATE Update $ T_{m} $.
			\STATE Update the local pheromones (optional).
		\ENDFOR 
	\ENDFOR
		\STATE  Update the global pheromones.
		\STATE Keep the best solution in this iteration if it improves the best in memory.
	\ENDWHILE
\RETURN The best solution built.
\end{algorithmic} 
\caption{ACO algorithm}\label{alg:Hormigas}
\end{algorithm}

\section{Description of the proposed ACO algorithm}
\label{SeccionACOParaClustering}
 
 The method starts by defining a list of $M$ artificial ants $ \mbf{h}_{1},\mbf{h}_{2},\ldots,\mbf{h}_{M}$, that will build a data clustering in $K$ classes (or clusters). At the beginning, it is possible to define the best ant in the colony, denoted by $ \mbf{h}^{*} $,  equal to $ \mbf{h}_{m} $ for some $ m=1,2,\ldots,M $, because in that moment there is no comparison parameter among them; thus the assignment could be random.

For ant $\mbf{h}_{m}$, with $ m=1,2,\ldots,M $, $K$ random points in the space of individuals 
(a hyperrectangle that contains all individuals) 
are considered, denoted by $ \mbf{g}_{1}^{m},\mbf{g}_{2}^{m},\ldots,\mbf{g}_{K}^{m} $. These points are interpreted as the initial centroids. $ C_{k}^{m} $ denotes the class $k$, with centroid  $ \mbf{g}_{k}^{m} $, which has been built by ant $m$. Also, $ \mbf{h}_{m} $ has a tabu list $ L_{m} $, which is a short term memory that contains the objects classified by $\mbf{h}_{m}$. In each iteration , in order to complete the tour, ant $m$  has to classify the objects not in $L_{m}$. When the iteration is done, all objects should be in $L_{m}$, this guarantees that the clustering process is complete.

During the clustering process, each ant randomly chooses an object that is not in its tabu list. Then, the ant should randomly select a class in which  to classify the object. If ant $m$ selects object $i$, then the process to choose the class uses a probabilistic roulette (see \cite{Talbi}). The probability that $\mbf{h}_{m} $ assigns object $i$ to class $ C_{k}^{m} $ is denoted by $ p_{ik}^{m} $. To calculate this probability it is necessary to consider the following factors:

\begin{itemize}

\item {\bf Visibility: } This factor is denoted by $ \eta_{ik}^{m} $, and it consists of the visibility of $\mbf{h}_{m}$, located on object $ \mbf{x}_{i} $, to ``see'' class $ C_{k}^{m} $. The visibility is defined as the reciprocal of the distance from object $ \mbf{x}_{i} $ to $\mbf{g}_{k}^{m} $, the centroid of class $C_{k}^{m}$. Thus, $ \eta_{ik}^{m}:=\tfrac{1}{d_{ik}^{m}} \text{, \ where }d_{ik}^{m}=d^{2}(\mbf{x}_{i},\mbf{g}^{m}_{k})=\norma{\mbf{x}_{i}-\mbf{g}_{k}^{m}}^{2}.$ If the visibility which $\mbf{h}_{m}$ has of class $C_{k}^{m}$  is large, then the probability of classifying $\mbf{x}_{i}$ in class $k$ is also large.

\item {\bf The pheromone trail: } The pheromone trail perceived by $\mbf{h}_{m}$ on the arc from $\mbf{x}_{i}$ to $ \mbf{g}_{k}^{m} $ is denoted by $ \tau_{ik} $. It quantifies pheromones that have been dropped by all ants which have classified the same object $\mbf{x}_{i}$ in its respective class $k$. If $ \tau_{ik} $ is large, then the probability of assigning class $k$ to cluster $\mbf{x}_{i}$ is going to increase.

\end{itemize}

Equation \eqref{ecu:Hormigas+Probabilidad} shows the formula used to calculate $ p_{ik}^{m}$, considering visibility and the pheromone trail, inspired by the corresponding formula used by the agent in the TSP:
\begin{equation}
\label{ecu:Hormigas+Probabilidad}
p_{ik}^{m}:=\dfrac{[\tau_{ik}]^{\alpha}\cdot [\eta_{ik}^{m}]^{\beta}}{\sum\limits_{r=1}^{K} [\tau_{ir}]^{\alpha}\cdot[\eta_{ir}^{m}]^{\beta} },
\end{equation}
where $ \alpha $ and $ \beta $ are parameters to be fitted.

On the other hand, when $ \mbf{h}_{m} $ chooses class $ C_{k}^{m} $ for object $ \mbf{x}_{i} $, the ant will register index $i$ in the respective tabu list $L_{m}$. Futhermore, $ \mbf{h}_{m} $ should do the following processes related to the assignment. 

\begin{itemize}
\item {\bf Local pheromone update:} Ant  $ \mbf{h}_{m} $ should drop a pheromone trail between object $ \mbf{x}_{i} $ and class $ C_{k}^{m} $. To do this, an auxiliary pheromone matrix was defined, denoted by $ \Gamma_{aux} $ with size $ n\times K $, such that entry $ ik $ of $ \Gamma_{aux} $ contains pheromones between $x_{i}$ and class $k$. This matrix has the format presented in Table \ref{tbl:MatrizFeromona}.

\begin{table}[H]
  \centering
$ \Gamma_{aux}= $\begin{tabular}{cccccc}
    \hline
            & $C_1$   & $C_2$   & $C_3$   & $\cdots$ & $C_K$ \\
    \hline
    $\mbf{x}_1$ &         &         &         &         &  \\
    $\mbf{x}_2$ &         &         &         &         &  \\
    $\mbf{x}_3$ &         &         &         &         &  \\
    $\vdots$ &         &         &         &         &  \\
    $\mbf{x}_n$ &         &         &         &         &  \\
    \hline
    \end{tabular}\medskip
    \caption{Auxiliary pheromone matrix.} \label{tbl:MatrizFeromona}%
\end{table}

Ant $ \mbf{h}_{m} $ will drop $ \Delta \tau_{ik}^{m} $ pheromones.  This quantity is defined by $\Delta\tau_{ik}^{m}:=\dfrac{Q}{d^{m}_{ik}},$
where $ Q $ is a parameter to be fitted. Finally, the local pheromone update is done by adding $ \Delta\tau_{ik}^{m} $ with the current entry $ik$  of $ \Gamma_{aux}$. 

\item {\bf Centroid update:} The final step in this process is to update the centroid $ \mbf{g}_{k}^{m} $ of class $C_k^m$. One possibility is using its definition $
\mbf{g}_{k}^{m}:=\tfrac{1}{\abs{C_{k}^{m}}}\dsum_{\mbf{x}\in C_{k}^{m}}\mbf{x}$. This option is not advisable because there are several unnecessary calculations. If fact, it is possible to update $ \mbf{g}_{k}^{m} $ recursively using its  value in the previous iteration in case object $\mbf{x}_{i}$ is transferred to class $C_k^m$. In \cite{Trejos:1998}  the following formula  is proven and is used to update the centroids more efficiently:
$\mbf{g}_{k}^{m}:=\tfrac{1}{\abs{C_{k}^{m}}}\left[ \left( \abs{C_{k}^{m}}-1 \right)\mbf{g}_{k}^{m}+\mbf{x}_{i} \right]$.
\end{itemize}

After each ant has clustered one object, it should randomly select a new object that is not in its tabu list. Next, the ant should follow the process previously described. This process is done $n$ times, clustering all objects by all ants.

When the process ends, each ant has a complete clustering of objects with the respective barycenters. Also, matrix $ \Gamma_{aux} $ contains pheromones that were dropped by ants. Entry $ik$ of $ \Gamma_{aux} $ contains pheromone $ \Delta \tau_{ik} $, which has been dropped by all ants that classified object $i$ in its respective class $k$. This quantity is represented by
$
\Delta \tau_{ik}= \dsum_{m=1}^{M}\Delta \tau_{ik}^{m} .
$

The next step is to calculate, for each ant, the within inertia. To do this, the classification done by each ant, and the respective barycenters, should be considered. Also, if one of the ants has a within inertia less than $ W(\mbf{h}^{*}) $ (the best inertia so far in memory), then $\mbf{h}^{*}$ (the best ant in memory) is required to be updated.

Global pheromones are stored in a  matrix $ \Gamma $ with the same structure as $ \Gamma_{aux}$. At the beginning, this matrix is initialized with values close to zero (indicating pheromone absence). When the travels of all ants finish, $ \Gamma $ is updated in entry $ik$ by 
 $
\Gamma_{ik}:=(1-\rho) \Gamma_{ik}+\rho \Delta\tau_{ik}, 
$
where $ \rho $ is the pheromone evaporation rate. 

When the pheromone updating process is done, matrix $ \Gamma_{aux} $ is initialized, to be used in the next iteration. Also, tabu lists (one per ant) are initialized, to start a new classification process.

As the final step to conclude the current iteration, an intensification process done by the best ant (the ant with lowest within inertia, denoted by $ \mbf{h}^{*} $) is developed. $ \mbf{h}^{*} $ repeats her path dropping extra pheromones in arcs it visited. The intensification follows the following rule:
\begin{equation*}
\Gamma_{ik}:=\sistema{\Gamma_{ik}+\tfrac{Q}{W(\mbf{h}^{*})}\;  \text{  if the object }i \text{ is in the class }k \text{ of }\mbf{h}^{*}, \\ \\
\Gamma_{ik} \;\;\;\;\;\;\;\;\;\;\;\;\;\;\;\;\text{otherwise;} }
\end{equation*} 
where $ W(\mbf{h}^{*}) $ denotes the within inertia of the classification done by $ \mbf{h}^{*} $. This ends the current iteration and a new clustering process is started, considering the following information: the global pheromone matrix $ \Gamma $, the barycenters of ants, which will be used as the initial centroids for the new classes, and the best ant $ \mbf{h}^{*} $.

Algorithm \ref{alg:HormigasImplementado} presents a detailed pseudocode of the BACOK. The K-means algorithm was applied (see line \ref{linHor:kMedias+1} in Algorithm \ref{alg:HormigasImplementado}) to each ant. The method is applied 
 after all ants have built their respective classifications, and 
 until the absolute difference between current inertia and previous inertia is less than 0.0001. Algorithm  \ref{alg:KMediasHormigas} shows how the local search strategy based on K-means works. If lines from \ref{linHor:kMedias+1} to \ref{linHor:kMedias+4} are eliminated from Algorithm \ref{alg:HormigasImplementado}, then BACO algorithm pseudocode is obtained.  Finally,  in the event that there has been no improvement, Algorithm  \ref{alg:HormigasImplementado}  uses  an iteration number (10 iterations) as stopping criterion  (see line  \ref{linHor:MetodoParo}). Consider that, $\Contador$ is increments in line \ref{lin:IncrementoContador}, but its value must be returned to zero every time a better solution (comparing with the best in memory) is found. This stopping criterion is based on the stagnation behavior concept presented in \cite{Dorigo+Maniezzo+Colorni}.

 \begin{algorithm}
 \begin{algorithmic}[1]
 \REQUIRE $ n $ (number of individuals), $ p $ (number of variables), $ K $ (number of clusters), $ M $ (number of ants),  and the parameters $ \alpha $, $ \beta $, $ Q $ and $ \rho $.
 \STATE  Build the initial colony with $m$ ants: $\mbf{h}_{1} $, $ \mbf{h}_{2},\ldots,\mbf{h}_{M} $.
 \STATE For each $ m=1,2,\ldots,M $ define $ L_{m}=\varnothing $, and randomly choose  $ \mbf{g}_{1}^{m},\ldots,\mbf{g}_{K}^{m} $. 
 \STATE  \Contador $ \leftarrow 0$
 \WHILE  {\Contador $ \;\leq  $ 10}  \label{linHor:MetodoParo}
 		\STATE $ \Contador \leftarrow \Contador +1 $ \label{lin:IncrementoContador}
 		\FOR{$ I:=1 $ \TO $ n $}
 		\FOR{$ m:=1 $ \TO $ M $}
 				\STATE Ant $ \mbf{h}_{m} $ chooses a random individual $ \mbf{x}_{i} $,  such that $i \notin L_{m} $. \medskip
 				\STATE Ant $\mbf{h}_{m} $ chooses $ k:=\textsf{\small Roulette}(p_{ik}^{m}) $, where
 				$ p_{ik}^{m}:=\tfrac{[\tau_{ik}]^{\alpha}\cdot [\eta_{ik}^{m}]^{\beta}}{\sum\limits_{r=1}^{K}[\tau_{ir}]^{\alpha}\cdot[\eta_{ir}^{m}]^{\beta}}.$
 				\STATE Individual $  \mbf{x}_{i} $ and index $i$ are assigned to $ C_{k}^{m} $ and $L_{m} $, respectively.
 				\STATE Let $ \left\langle \Gamma_{aux} \right\rangle_{ik}:= \left\langle \Gamma_{aux} \right\rangle_{ik}+\Delta\tau_{ik}^{m}$,  where  $\Delta\tau_{ik}^{m}=\frac{Q}{d^{m}_{ik}} $.
 				\STATE Let $ \mbf{g}_{k}^{m}:=\frac{1}{\abs{C_{k}^{m}}}\left[ \left( \abs{C_{k}^{m}}-1 \right)\mbf{g}_{k}^{m}+\mbf{x}_{i} \right] $.
		\ENDFOR
		\ENDFOR
		\STATE Let $ \mbf{h}^{*}:= \textsf{\footnotesize BestAnt}(\mbf{h}_{1},\ldots,\mbf{h}_{M},\mbf{h}^{*}) $.
 		\STATE For $i=1,\dots,n$ and $k=1,\dots,K$ let $ \left\langle \Gamma\right\rangle _{ik}:=\tau_{ik} $,\newline where $ \tau_{ik}:=(1-\rho)\ \left\langle \Gamma \right\rangle_{ik} +\rho  \left\langle \Gamma_{aux} \right\rangle _{ik}$.
 		\STATE Intensify the best trail. For all $i(i=1,\dots,n)$, if individual $i$ in $\mbf{h}^*$ was classified in cluster $k$ do $ \left\langle \Gamma \right\rangle_{ik}=\left\langle \Gamma \right\rangle_{ik}+{Q}/{W(\mbf{h}^{*})}$
 	
 				\STATE If the inertia of $\mbf{h}^*$ improves the best inertia keeped in memory, reset \Contador\hspace{-0.1cm}.
 				\FOR{$ m:=1 $ \TO $ M $} \label{linHor:kMedias+1}
 						\STATE Apply K-means to $ \mbf{h}_{m} $. \label{linHor:kMedias+2}
 						\STATE Update $ \mbf{h}^{*} $ if there was an improvement from the K-means application. \label{linHor:kMedias+3}
 				\ENDFOR \label{linHor:kMedias+4}
 	\ENDWHILE  
 \RETURN $ \mathbf{h^{*}} $
 \end{algorithmic} 
 \caption{BACOK algorithm.}\label{alg:HormigasImplementado}
 \end{algorithm}

 \begin{algorithm}
 \begin{algorithmic}[1]
 \REQUIRE One ant $ \mbf{h}$.
 \STATE  {\small \sf PreviousInertia}$ \leftarrow -1$.
 \WHILE  {$ \abs{\text{{\small \sf PreviousInertia}}-W(\mbf{h}_{m})} >0.001$}           
 		\STATE  {\small \sf PreviousInertia}$ \leftarrow  W(\mbf{h}_{m})$
 		\STATE For $ \mbf{h}$, build clusters $ C_{1},C_{2},\ldots,C_{K} $, using barycenters $ \mbf{g}_{1},\ldots,\mbf{g}_{K} $. To do that, assign each individual $ \mbf{x}_{i} $ to the class with its barycenter closest to $ \mbf{x}_{i} $.
 		\STATE Recalculate the barycenters $\mbf{g}_{1},\mbf{g}_{2},\ldots,\mbf{g}_{K} $ with:\newline
 		$ \mbf{g}_{k}=\dfrac{1}{\abs{C_{k}}}\dsum_{\mbf{x}_{i}\in C_{k}}\mbf{x}_{i} \text{, for all } k=1,2,\ldots,K.$
 	\ENDWHILE  
 \RETURN A new ant $ \widehat{\mbf{h}} $.
 \end{algorithmic} 
 \caption{Local search strategy based on K-means applied in BACO.}\label{alg:KMediasHormigas}
 \end{algorithm}

\section{Parameter analysis}
\label{SeccionExperimental}

To develop the parameter analysis three data tables (T105, T525 and T2100) were built, with randomly generated normal variables. The data sets T105 ($n=105$ and $p=6$) and T525 ($n=525$ and $p=6$) consists of 105 and 525 objects, respectively. Both sets have seven clusters ($K=7$), such that six classes have variance equal to $\sigma^2=1$, and the seventh class has $\sigma^2=3$. The data set T105 has a ``big'' class with 51 objects, and the remaining six groups with 9 objects. Meanwhile, T525 has a class with 265 objects, and the remaining objects are equitably distributed in the other groups. The $W(P)$ reference values for T105 and T525 were calculated using the equation (\ref{W}), thereby $7.62467183$ and $7.45610263$ were obtained for these tables, respectively. Table T2100 has 2100 objects, seven clusters with the same cardinality and all classes have different variances. The $W(P)$ reference value for this set is $22.56959210$.   

The Algorithm \ref{alg:HormigasImplementado} has four parameters that should be fitted, with the aim of achieving good performance. Parameters $ \alpha $ and $ \beta $ control the relative weights assigned to pheromone concentration and ant visibility, respectively. Meanwhile, $ \rho $ represents the pheromone evaporation rate, used to update the pheromone matrix. Finally, parameter $Q$ is a \textit{pheromone amplification constant}. 

To develop the parameter analysis tables T105 and T525 were used, and for each table, and for each parameter combination, 200 multistart runs were done. Based on the ranges presented in \cite{Dorigo+Maniezzo+Colorni}, in the current experiment a further analysis was developed, using
$\rho\in \{0.1,0.2,\ldots,0.9\}$, $\alpha, \beta \in \{0,0.5,1,1.5,\ldots, 6\}$, and $Q\in \{50,100,150,\ldots,500\}$. In total $9\times 13\times 13 \times 10=15210$ combinations were run for each table. This analysis used $M=10$ (the number of ants).

\begin{figure}
\centering
\subfigure[Contour map for $ \rho=0.1 $ and $ Q=50 $. ]{\includegraphics[scale=0.35]{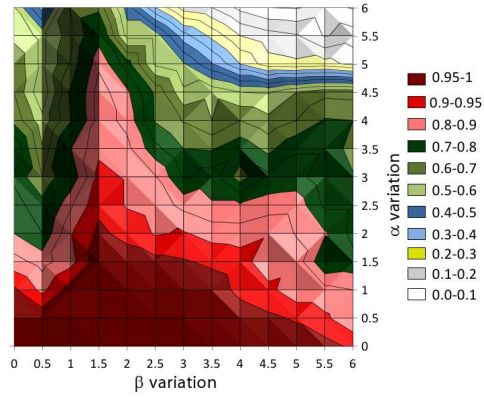}\label{fig:OptiOCH105rho01Q50}} \hspace{0.2cm}
\subfigure[Contour map for $ \rho=0.5 $ and $ Q=50 $. ]{\includegraphics[scale=0.35]{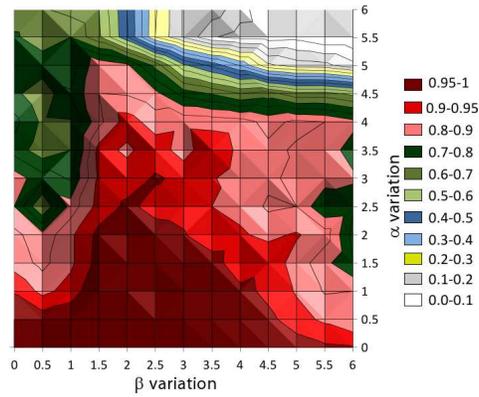}\label{fig:OptiOCH105rho05Q50}}

\subfigure[Contour map for $ \rho=0.9 $ and $ Q=50 $. ]{\includegraphics[scale=0.35]{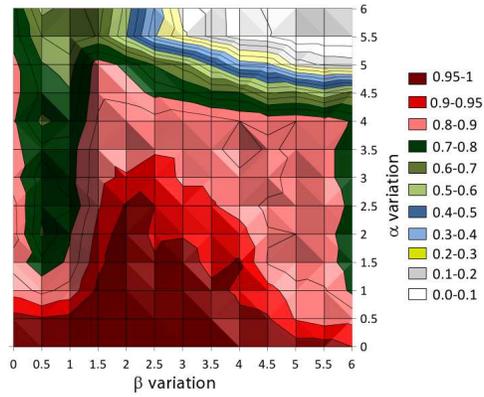}\label{fig:OptiOCH105rho09Q50}}
 \hspace{0.2cm}

\caption{Some examples of contour maps created with the performance percentages, for $Q=50$, $\rho=0.1,0.5,0.9$, and variants values for $\alpha$ and $\beta$. Analysis done with table T105.}\label{fig:OptiOCH105}
\end{figure}

The pictures in Figure  \ref{fig:OptiOCH105} show some examples of the 90 contour maps built with the performance percentages (each percentage represents how many times the algorithm scores the $W(P)$ reference value, in the 200 runs) obtained with table T105, for the different parameter combinations. For example, Figure \ref{fig:OptiOCH105rho01Q50} shows the contour map for $\rho=0.1$, $Q=50$ and $\alpha,\beta\in \{0,0.5,1,1.5,\ldots, 6\}$. This analysis showed that $\rho=0.5$ was the best option, because the best performance zone for $\rho=0.5$ (the darker red zone in Figure \ref{fig:OptiOCH105rho05Q50}) is better (largest area) than those of the remaining $\rho$ values.

On the other hand, very similar contour maps were obtained when $\rho$ was fixed, and $Q$  varied from $50$ to $500$ (10 contour maps per each $\rho$ value). This showed evidence that $Q$ was not an important parameter in this experiment. And this coincides with the observation presented in \cite{Dorigo+Maniezzo+Colorni}, which indicates that $Q$ has a negligible influence in the algorithm. Therefore, the parameter $Q$ was fixed at $250$ (the range middle value), but also could be fixed at $100$, as they did.   

\begin{figure}[h]
\centering
\subfigure[Results obtained with table T105. ]{\includegraphics[scale=0.40]{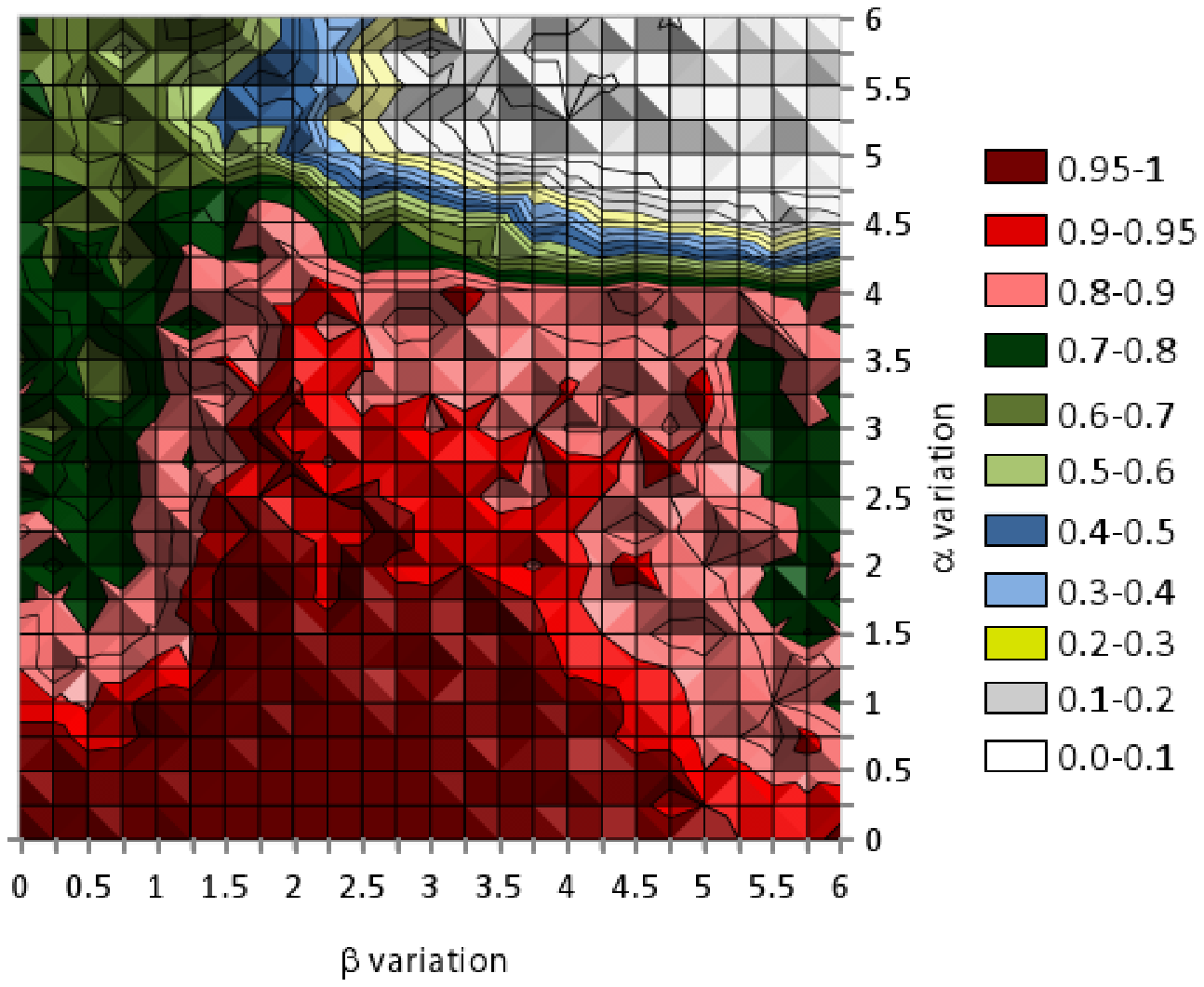}\label{fig:MallaAlphaBeta6x6:105}} 
\hspace{0.2cm}
\subfigure[Results obtained with table T525. ]{\includegraphics[scale=0.40]{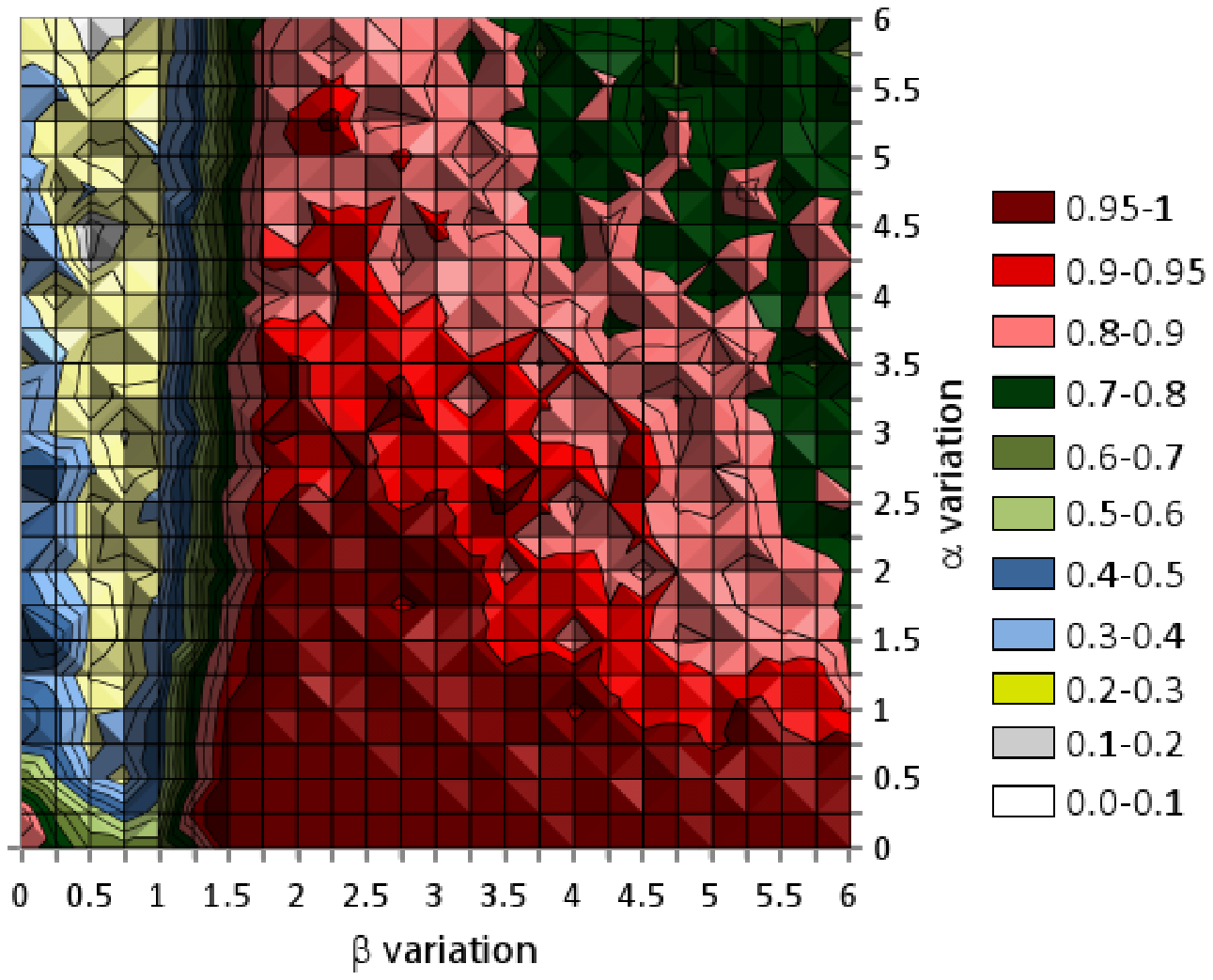}\label{fig:MallaAlphaBeta6x6:525}}

\caption{Contour maps created with the performance percentages, with the fixed values $\rho=0.5$ and $Q=250$.}\label{fig:MallaAlphaBeta6x6}
\end{figure}

\begin{figure}[h]
\centering
\subfigure{\includegraphics[scale=0.6]{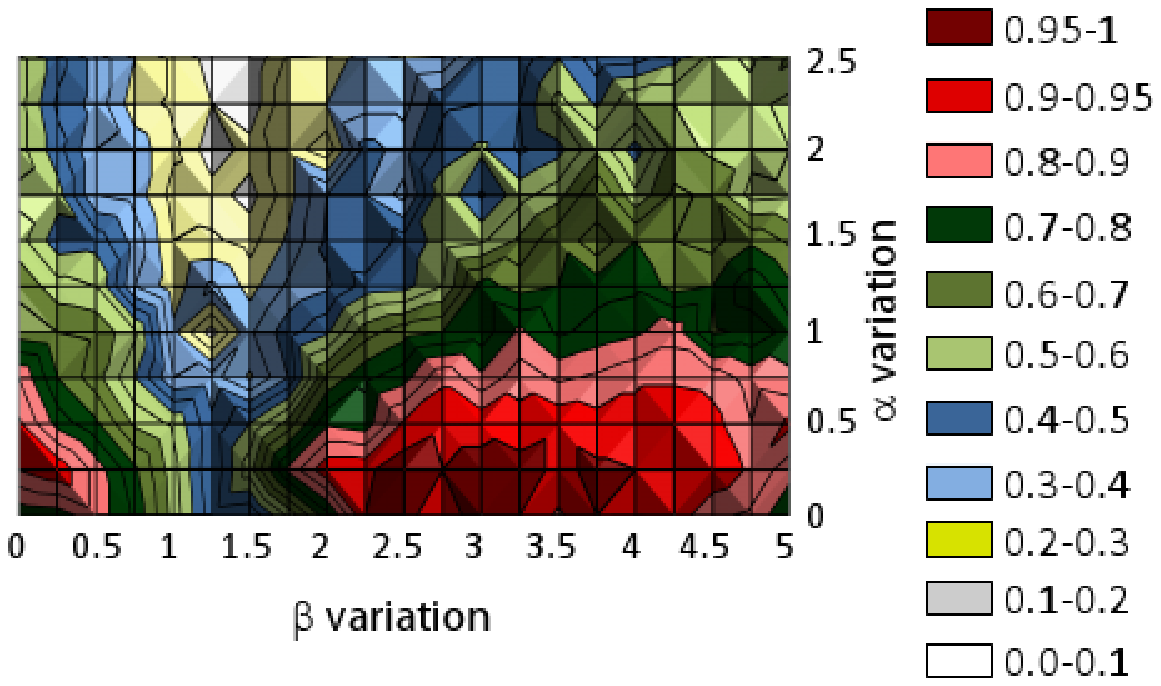}}
\caption{Contour maps created with the performance percentages, with the fixed values $\rho=0.5$ and $Q=250$, in table T2100.}\label{fig:MallaAlphaBeta2100}
\end{figure}

Next, an analysis for $\alpha$ and $\beta$ was developed with tables T105 and T525, using $\rho=0.5$, $Q=250$, and $\alpha,\beta \in \{0,0.25,0.5,0.75,\ldots,6\}$. Figure \ref{fig:MallaAlphaBeta6x6} shows the contour maps obtained in this process. This analysis was not enough to determine optimum values for $\alpha$ and $\beta$. Figures \ref{fig:MallaAlphaBeta6x6:105} and \ref{fig:MallaAlphaBeta6x6:525} only suggest that the best performance is probably obtained when $1.5\leq \beta\leq 5$ and $0<\alpha\leq 2.5$. For this reason, an extra analysis was developed with table T2100. Figure \ref{fig:MallaAlphaBeta2100} shows that any combination for $\alpha$ and $\beta$ in the dark red region could be taken. Therefore, for this experiment the combination $\beta=2.5$ and $\alpha=0.25$ was selected. Summarizing, the parameters were chosen as $\alpha=0.25$, $\beta=2.5$, $\rho=0.5$ and $Q=250$.    

\section{Extra data sets, results and discussion}
\label{SeccionDiscusionResultados}

\begin{figure}[hbt]
	\centering
	\subfigure[A1. ]{\includegraphics[scale=0.65]{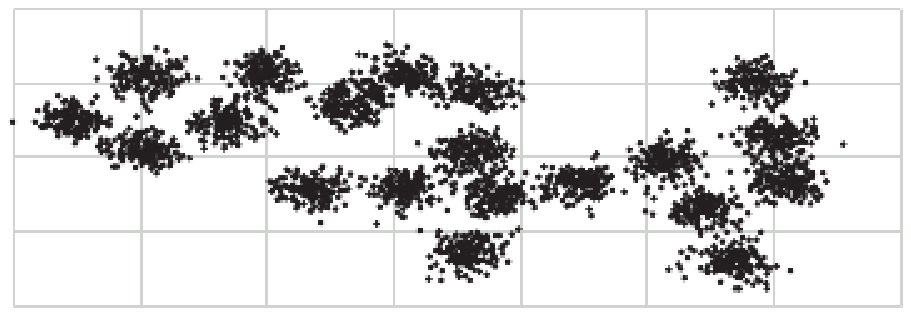}\label{fig:DataSetA1}} 
	\hspace{0.2cm}
	\subfigure[A2.]{\includegraphics[scale=0.65]{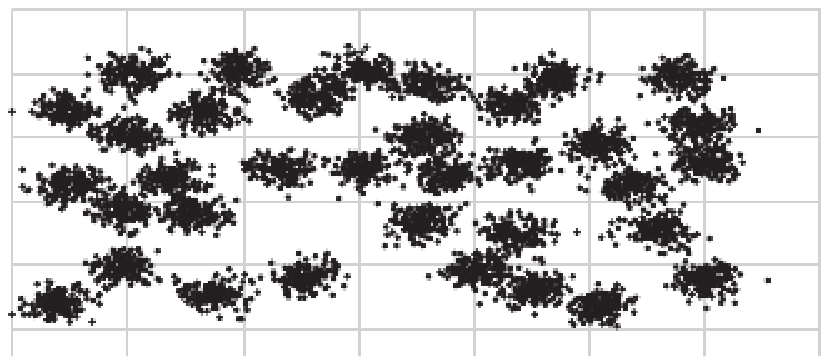}\label{fig:DataSetA2}}
	\subfigure[A3. ]{\includegraphics[scale=0.65]{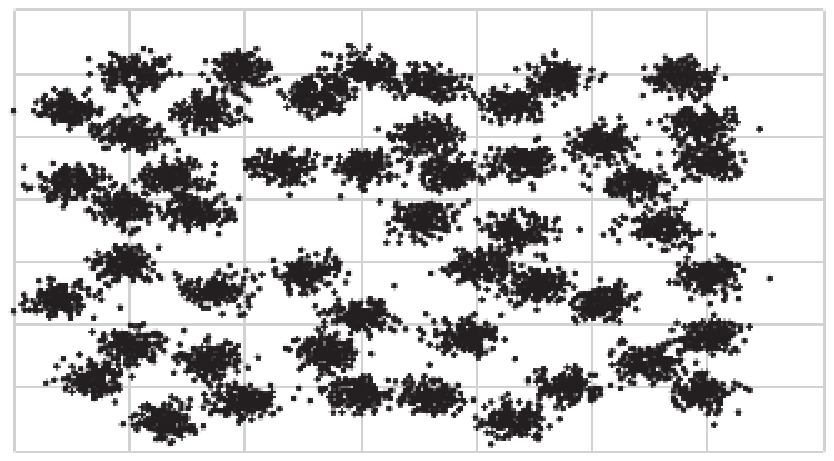}\label{fig:DataSetA3}} 
	\hspace{0.2cm}
	\subfigure[S1.]{\includegraphics[scale=0.65]{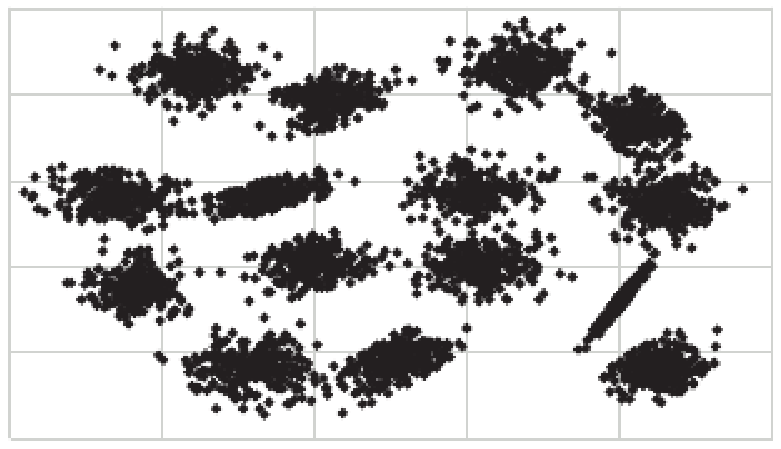}\label{fig:DataSetS1}}
	\subfigure[S2. ]{\includegraphics[scale=0.65]{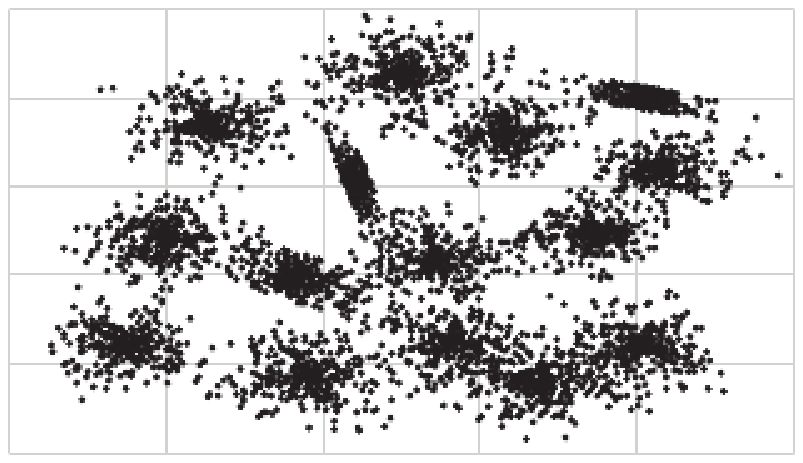}\label{fig:DataSetS2}} 
	\hspace{0.2cm}
	\subfigure[S3.]{\includegraphics[scale=0.65]{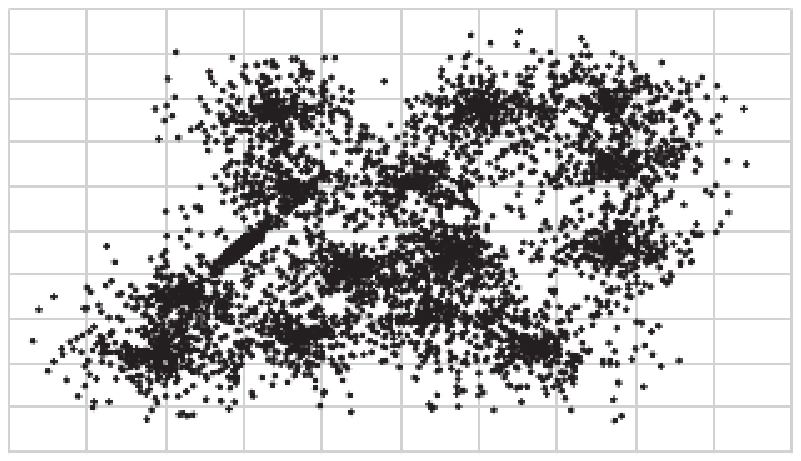}\label{fig:DataSetS3}}
	\subfigure[S4.]{\includegraphics[scale=0.65]{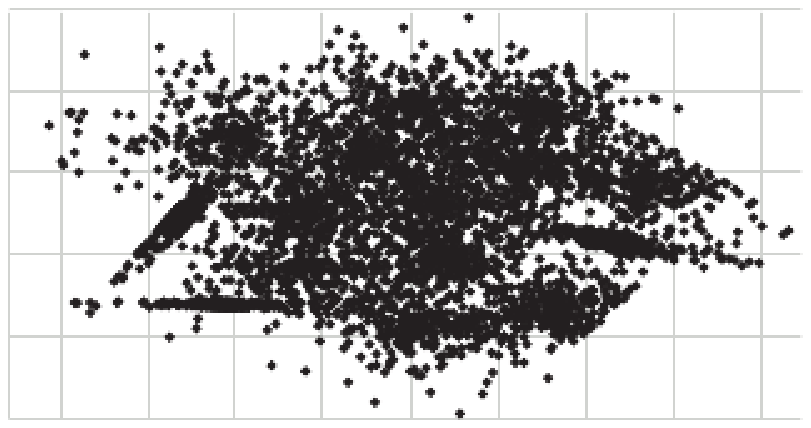}\label{fig:DataSetS4}}	
	\caption{Two-dimensional representation for the datasets on groups A and B.}\label{fig:DataSetsAandS}
\end{figure}

A personal computer with 8GB of RAM memory and an Intel Core i7-4712MQ CPU@2.30GHz processor, was used in this experiment. In order to develop a comparison among the algorithms BACO, BACOK and KM, five real life data sets were downloaded from the website of UCI repository of machine learning databases \cite{UCI}: iris ($n=150$, $K=3$ and $p=4$), wine ($n=178$, $K=3$ and $p=13$), glass identification ($n=214$, $K=6$, $p=9$), red-wine quality ($n=1599$, $K=3$, $p=11$) and white-wine quality ($n=4898$, $K=3$, $p=11$) data sets. In \textit{glass} data set the first attribute was not considered as a variable, because it is an identification number (for this reason $p=9$). Furthermore, $K$ was fixed at $6$ because the type of glass number 4 is not present in this data set (in total, there are 7 types of glass). In \textit{wine quality} (both tables), the attribute number 12 was not considered because it is an output variable. Additionally, two groups (A and S) of bidimensional synthetic data sets were considered (downloaded from \cite{TablesAS}), which are described on \cite{Franti+Sieranoja}. Group A (3 sets) varies the number of clusters, and the group S varies the overlaping among the clusters (4 sets). All cases use $p=2$. Table \ref{tbl:SetAS} summarizes the main features of these sets and Figure \ref{fig:DataSetsAandS} shows a bidimensional representation for each set. Also, the \textit{ground truth} centroids for these data sets are available on \cite{TablesAS}, hence it was possible to analyze if the proposed BACOK algorithm was generating a reasonable clustering for the data. The \textit{centroid index} (CI) presented in \cite{Franti+Rezaei+Zhao} is a cluster level similarity measure, based on the cluster centroids, which can be used to compare one clustering against other solution or the ground truth, if is available. The algorithm BACOK was executed 100 times on sets A1, A2, A3, S1, S2, S3, and S4, and the best solution found, in each case, was compared with the ground truth solution, using the CI value. In all cases, the CI value was equal to zero, therefore according to  \cite{Franti+Sieranoja}, our algorithm is properly clustering those datasets. This experiment was made with 20 ants ($M=20$)  and the parameters $\alpha=0.25$, $\beta=2.5$, $\rho=0.5$, and $Q=250$. Finally, using for each set the centroids of the best solution and the definition of $W(P)$ (see equation \ref{W}), the best within inertia for each set ($W_{best}$) was calculated (see column number 4 on Table \ref{tbl:SetAS}). 

Table \ref{tbl:Sumarize} 
	summarizes the results obtained with the three algorithms. The performance of each algorithm is represented by a percentage, and this corresponds the number of times which the algorithm scored the $W_{best}$ value in 100 multistart runs. The algorithm BACO also used $M=20$, $\alpha=0.25$, $\beta=2.5$, $\rho=0.5$, and $Q=250$. Meanwhile, the KM algorithm iterates until the difference between two consecutive within inertias is less than $0.001$. The symbol ``-'' used in Table \ref{tbl:Sumarize} means the algorithm did not attend the $W_{best}$ reference value in any of the 100 runs. Also, the standard deviation of inertia, the average time and the standard deviation of time, in those 100 executions, are presented on Table \ref{tbl:Sumarize}.

\begin{table}[htb]
\centering
\scalebox{0.9}[0.9]{
\begin{tabular}{cccc}
\hline
\text{Data set}& $n$ & $K$ & $W(P)$ reference value\\
\hline
A1        & 3000 &  20 & 4048752.50 \\
A2        & 5250 &  35 & 3864140.31 \\
A3        & 7500 &  50 & 3858322.01 \\
S1        & 5000 &  15 & 1783523123.37 \\
S2        & 5000 &  15 & 2655821898,14 \\
S3        & 5000 &  15 & 337791436.87 \\
S4        & 5000 &  15 & 3140628447.25 \\
\hline
\end{tabular}}  \hspace{1cm}
\caption{Main features for sets on group A ans S.} \label{tbl:SetAS}
\end{table}
\newpage

\pagebreak

\begin{table}[htb]
	\centering
	\begin{turn}{90}
	\scalebox{0.85}[0.85]{
		\begin{tabular}{c|c|ccc|ccc}
			\hline
			&\multicolumn{1}{|c}{}&\multicolumn{3}{|c}{Performance - Standard deviation of inertia}&\multicolumn{3}{|c}{Average time - Standard deviation of time (seconds) }\\ 
			\text{Data set}& \text{W$_{best}$}  & BACO    & KM & BACOK & BACO    & KM & BACOK \\
			\hline
			Iris         & 0.52136 & 34\%$\;$-$\;$0.00104 & 5\%$\;$-$\;$0.19743 & 100\%$\;$-$\;$0 &0.16467 $\;$-$\;$0.05200   &0.00027 $\;$-$\;$0.00008  &0.13550 $\;$-$\;$0.02911\\
			Wine         & 13318.48 & 8\%$\;$-$\;$41.3007 & 100\%$\;$-$\;$ 0 & 100\%$\;$-$\;$0 &0.25580 $\;$-$\;$0.08328  &0.00048 $\;$-$\;$0.00004  &0.31380 $\;$-$\;$0.04476\\
			Glass        & 1.570377 & $\;$-$\;$ & $\;$-$\;$ & 93\%$\;$-$\;$0.0001 &$\;$-$\;$  &$\;$-$\;$  &0.64427 $\;$-$\;$0.17442\\
			Red-Wine Q   & 247.2075 & $\;$-$\;$ & 1\%$\;$-$\;$0.00717  & 100\%$\;$-$\;$0  &$\;$-$\;$  &0.00794$\;$-$\;$0.00072  &3.28075$\;$-$\;$0.25986\\
			White-Wine Q & 560.4186 & $\;$-$\;$& 83\%$\;$-$\;$0.00022 &100\%$\;$-$\;$0  &$\;$-$\;$  &0.03349 $\;$-$\;$0.00778  &16.59486 $\;$-$\;$2.48081\\
			A1 & 4048752.50 & $\;$-$\;$& $\;$-$\;$ &80\%$\;$-$\;$244433.51  &$\;$-$\;$  &$\;$-$\;$  &14.84363 $\;$-$\;$4.49093\\
			A2 & 3864140.31 & $\;$-$\;$& $\;$-$\;$ &90\%$\;$-$\;$70488.07  &$\;$-$\;$  & $\;$-$\;$  &62.97577 $\;$-$\;$15.71993\\
			A3 & 3858322.01 & $\;$-$\;$& $\;$-$\;$ &50\%$\;$-$\;$139284.45 & $\;$-$\;$  & $\;$-$\;$  &164.00306 $\;$-$\;$54.02449\\
			S1 & 1783523123.37 & $\;$-$\;$& $\;$-$\;$ &100\%$\;$-$\;$0  &$\;$-$\;$  & $\;$-$\;$  &11.94175 $\;$-$\;$3.86356\\
			S2 & 2655821898.14 & $\;$-$\;$& $\;$-$\;$ &100\%$\;$-$\;$0  &$\;$-$\;$  & $\;$-$\;$  &12.65522 $\;$-$\;$0.93671\\
			S3 & 3377914369.87 & $\;$-$\;$& $\;$-$\;$ &87\%$\;$-$\;$1390.99  &$\;$-$\;$  & $\;$-$\;$  &22.69879 $\;$-$\;$5.31946\\
			S4 & 3140628447.25 & $\;$-$\;$& $\;$-$\;$ &10\%$\;$-$\;$7221.77  &$\;$-$\;$  & $\;$-$\;$  &30.60434 $\;$-$\;$8.88218\\
			\hline
	\end{tabular}
	}
\end{turn} 
 \hspace{1cm}
	\caption{Performance comparison among BACO, BACOK and KM.} 
	\label{tbl:Sumarize}
\end{table}

\clearpage

Table \ref{tbl:Sumarize} shows how the algorithm BACOK performed very good on the available data sets. This final comparison is valuable because it
reinforces one of the principal contributions of this paper: the BACO and KM algorithms  did not show good results in most all data sets, but our algorithm uses the potential of K-means to improve the algorithm BACO, and then significantly better results were obtained. That 
hybridization process presented in algorithm BACOK reveals how the K-means algorithm itself could not work well, but it can be used to improve other heuristic algorithms. Finally, the lowest performance reported by algorithm BACOK was in the set S4, which has the highest level of overlap (see Figure \ref{fig:DataSetsAandS}).

\section{Conclusions}
\label{SeccionConclusiones}

We have presented a hybrid clustering method based on the ant colony optimization metaheuristic and the K-means algorithm. The method is based on some features developed for ACO in the traveling salesman problem and it is improved by the K-means algorithm in each iteration.
The adaptation to the clustering problem takes into account the representation of clusters by barycenters, and therefore the distance between objects and barycenters is used for defining visibility and the pheromone trail.

After a  extensive parameter fitting, an experimentation was implemented in order to evaluate the method. It performed very well, attaining the reference value for the inertia in each data table, in reasonable time. Furthermore, the method showed very good results when it was applied to other benchmark data sets, where the ground truth for each set was available.   

Finally, the experiment revealed the parameter $Q$ does not have a relevant role in the ACO algorithm, but the algorithm is very sensitive to the values assigned to the parameters $\alpha$, $\beta$ and $\rho$. The parameter fitting process was necessary to improve the algorithm performance and it gave the combination $\alpha=0.25$, $\beta=2.5$ and $\rho=0.5$.

\subsection*{Acknowledgments}

J. Chavarr\'ia and J.J. Fallas were supported with project 5402-1440-3901 of the Research Vice Rectory, Costa Rica Institute of Technology.
J. Trejos was supported with project 821-B1-122 of the Research Vice-Rectory, University of Costa Rica.


\end{document}